# A Large-Language Model Framework for Relative Timeline Extraction from PubMed Case Reports

Jing Wang, Ph.D.[1], Jeremy C. Weiss, M.D., Ph.D.[1]
[1]National Library of Medicine, Bethesda, Maryland, USA

**Abstract**

*Timing of clinical events is central to characterization of patient trajectories, enabling analyses such as process tracing, forecasting, and causal reasoning. However, structured electronic health records capture few data elements critical to these tasks, while clinical reports lack temporal localization of events in structured form. We present a system that transforms case reports into textual time series—structured pairs of textual events and timestamps. We contrast manual and large language model (LLM) annotations (n=320 and n=390 respectively) of ten randomly-sampled PubMed open-access (PMOA) case reports (N=152,974) and assess inter-LLM agreement (n=3,103 N=93). We find that the LLM models have moderate event recall (O1-preview: 0.80) but high temporal concordance among identified events (O1-preview: 0.95). By establishing the task, annotation, and assessment systems, and by demonstrating high concordance, this work may serve as a benchmark for leveraging the PMOA corpus for temporal analytics. Code is available at: https://github.com/jcweiss2/LLM-Timeline-PMOA/.*

**Introduction**

Clinical event timelines are key analytic devices in use in areas ranging from the visualization of patient trajectories in electronic health records to the development and update of clinical practice guidelines. While many automated documentation systems capture structured health information in relational databases with timestamping, textual data modalities often lack temporal granularity beyond creation and submission dates, despite being the main tool care providers use to document and communicate patient history and care planning. For example, case reports and other clinical notes may contain sections such as past medical history and history of present illness where the times of health events are mentioned. The time of these events relative to the case presentation time can then be inferred, *e.g.*, "6 months prior", "5-day history of fever and chills", or "history of diabetes diagnosed in January 2024". We argue that a gap is present in our temporal understanding of patient trajectories that can be bridged by improving temporal richness of textual information, or conversely by improving timestamped event capture with text.

Several lines of clinical informatics research have addressed this problem with tasks including temporal relation extraction and formal temporal specifications[1-4]. These investigations culminated in the i2b2 2012 temporal relations challenge, where clinical events and time expressions were extracted and temporal relations were annotated from 310 hospital discharge summaries[4]. A series of models have shown iterated improvement using natural language processing and BERT-based models[5], and recently with large language models (LLMs)[6]. However, while temporal relations capture event ordering in time, they do not capture timestamp, so multimodal alignments, *e.g.*, with structured time-stamped data elements, and tasks such as time-to-event analysis cannot be performed without additional annotation effort. To address this concern, Leeuwenberg and Moens and Frattallone et al. focused on time interval annotations, which showed that multimodal (text-and-tabular) deep learning and BERT-based models could be used to predict these intervals with success[7,8]. When the open-source large language model LLaMa-2-13B was applied, it did not perform competitively on the interval prediction[8]. An open question remains about the performance of the highest performing LLMs closed-source models, which remains unverified due to the challenges of using those models on private and semi-public clinical data resources.

Meanwhile, recent improvements in performance on benchmarks in clinical question answering[9] suggest that LLMs may be able to help on other reasoning tasks in clinical informatics, one of them being timeline construction. Thus, we seek to investigate the performance of high-performing LLM models on a public repository, PubMed Open Access case reports[10]. Case reports have thematic overlap with the discharge summaries analyzed in the i2b2 corpus, as many case reports are written to describe interesting clinical findings and to serve as communication artifacts for discussing clinical reasoning and best practices. As we show, our extraction identifies 152,974 case reports, and among a random sample of one hundred, 93 are identified as single-case case reports, suggesting the PMOA could be a rich source for clinical timeline analysis with sufficient sample size to explore medical care broadly and subdomains of interest.

To investigate feasibility of this effort, our work characterizes an LLM framework, describing the case selection, annotation process, and assessment. To validate the LLM responses, we manually annotated a small random subset of case reports and contrasted our results with inter-LLM agreement performance on a larger subset (akin to inter-annotator agreement). The results are promising, yet work remains, and we intend for our exposition to aid in considering system design choices and assessments as the effort to construct clinical timelines from free text continues to be refined.

*Related Work*
Our LLM annotation pipeline involves steps of event extraction and temporal assignment relative to case presentation, and both tasks have been addressed in prior work. LLMs have been used in clinical domains for state of the art question answering[9], entity extraction[11], and synthesis[12]. Among those focused on temporal extraction and modeling, many have used the i2b2 temporal relations challenge of 2012, which provided annotations of hospital discharge summaries, specifically clinical named entities and their temporal ordering. These include Kougia et al's comparison of GPT 3.5, Mixtral, and PMC-LLaMa, and others, which showed competitive performance to multimodal BERT-based models for clinical temporal relation extraction[6,13]. For non-clinical temporal relations, GPT 4.0 zero-shot extraction has been applied, however in that study, BERT-based strategies outperformed the prompt-engineered methods[14].

The temporal information of our focus is the time relative to case presentation, which is closer to works that use models to annotate or extract clinical timelines[7,8]. In these works, events with associated time intervals are modeled probabilistically, whereas our work focuses on events associated with a single relative time for simplicity of the LLM request. Leeuwenberg provided baseline LSTM models for timeline prediction, and Cheng and Weiss demonstrated improved performance with BERT-based models with clinical tagging elements[5,7]. In a comparison of timeline annotations of i2b2 discharge summaries conducted by a clinical expert, Frattallone et al. showed that multi-modal BERT-based models outperformed zero-shot Llama-2-13b extraction[8]. In our case of PubMed Open Access case reports, an additional data modality, *e.g.*, structured tabular data from electronic health records, is unavailable, and the multimodal augmented BERT-based strategies would be difficult to apply. Finally, the PubMed Open Access corpus has also been used for other downstream modeling tasks, with several successes such as in retrieving biomedical concepts[15] and serving pre-trained biomedical BERT models[16].

*Contributions*
Our work presents the following contributions. First, we introduce a novel framework for characterizing and comparing relative timelines extracted from case reports. Second, we apply our framework to the PubMed Open Access corpus, identifying case reports eligible for event-time extraction, generating structured event-time outputs, and assessing the quality and agreement of the responses. To our knowledge, this represents the first use of PubMed case reports for timeline recovery. Third, our results suggest that clinical time construction appears feasible using LLMs, particularly with respect to providing temporal ordering, yet with work remaining to improve completeness and relative timing. Finally, we provide our annotations as a pilot corpus for benchmarking.

**Methods**
Our annotation system comprised the following parts: extraction, annotation, and assessment (Figure 1), to facilitate comparison of expert annotation with LLM annotations and inter-LLM agreement.

*Extraction.* The text corpus used came from the PubMed Open Access (PMOA) repository as of 2024/04/18 comprising 1,488,346 full-text articles. While PubMed has metadata to indicate case reports, we found the 'case report' labeling had low sensitivity and low specificity upon manual review. Instead, we used the following regular expressions to identify clinical case reports: (i) presence of 'case report' or 'case presentation', and (ii) presence of the string 'year-old' or 'year old', all ignoring capitalization, in the body of the article. To find the body of the article we retained only the text between '==== BODY' and '==== Refs', which are section separators encoded in the PMOA corpus. This approach led to the identification of 152,974 probable case reports. To verify the precision of our case report identification approach, we took the matching articles and randomly sampled 100 of them for manual inspection. Of the 100 case reports, one had no case presentation, and six had multiple cases. We selected the remaining 93 articles for further analysis. For parameter and design selection—*e.g.*, prompt tuning and textual distance—we randomly sampled an additional non-overlapping 5 case reports for use as a development set.

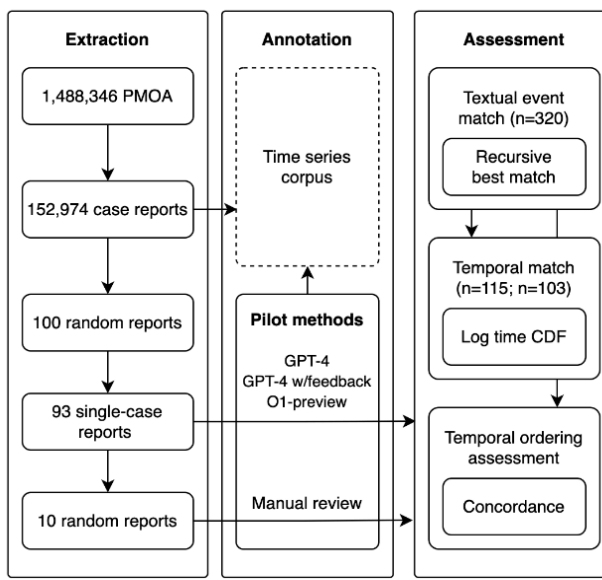

Figure 1. Flow diagram of the relative time annotation system

*Annotation*.

Our annotations consisted of LLM annotations of the 93 case reports and among those, we selected 10 case reports randomly for manual annotation by a clinically-trained expert. The annotation process was to extract events as defined by the ISO-TimeML specification[3], which informally correspond to events that can be inferred to have happened or hold true. For manual annotation, the expert was asked to select text spans without modification as events, with two exceptions: (i) allowing expanding conjunctive lists into its constituent elements, and (ii) allowing the attachment of the prefix of "history of" to the text span. These exceptions were meant to increase the usability of the events, while also maintaining the inverse mapping to the original text position so that the text span context could be used in downstream modeling. Once an event was specified, the annotator was asked to assign a time relative to the case presentation time, *e.g.*, admission time or encounter date. While other temporal frameworks have assigned time intervals to events[7,8], during our initial exploration with LLM time annotations we found the capture of intervals often resulted in nonsensical output formats or times. If the text described an interval corresponding to an event, the annotator was asked to use the start time, *i.e.*, the lower bound of the time interval. The annotation process resulted in two column files (event and time) with 320 events.

For LLM annotations, we used GPT-4 (gpt-4-0613) and O1-preview (o1-preview-2024-09-12), as they produced better results than Llama-3-8B-Instruct, Llama-3-70B, and GPT-4o in our initial exploration. We conducted an informal, manual search over prompts by inspecting results on the development set and settled on a strategy incorporating principles of few-shot prompting and incorporating feedback. We compared three strategies, GPT-4 without feedback ("GPT-4"), GPT-4 with response and feedback ("GPT-4 w/feedback"), and O1-preview without feedback ("O1-preview"). Our final prompt was of the form:

"{base_prompt}\n\nOriginal Text: {original_text}\n\nUpdates: {response}{feedback_prompt}",

where the bracketed placeholders are replaced with the corresponding string values. We used the {base_prompt}:

"You are a physician. Extract the clinical events and the related time stamp from the case report. The admission event has timestamp 0. If the event is not available, we treat the event, e.g. current main clinical diagnosis or treatment with timestamp 0. The events that happened before event with 0 timestamp have negative time, the ones after the event with 0 timestamp have positive time. The timestamp are in hours. The unit will be omitted when output the result. If there is no temporal information of the event, please use your knowledge and events with temporal expression before and after the events to provide an approximation. We want to predict the future events given the events happened in history. For example, here is the case report.

An 18-year-old male was admitted to the hospital with a 3-day history of fever and rash. Four weeks ago, he was diagnosed with acne and received the treatment with minocycline, 100 mg daily, for 3 weeks. With increased WBC count, eosinophilia, and systemic involvement, this patient was diagnosed with DRESS syndrome. The fever and rash persisted through admission, and diffuse erythematous or maculopapular eruption with pruritus was present. One day later the patient was discharged.

Let's find the locations of event in the case report, it shows that four weeks ago of fever and rash, four weeks ago, he was diagnosed with acne and receive treatment. So the event of fever and rash happened four weeks ago, 672 hours, it is before admitted to the hospital, so the time stamp is -672. diffuse erythematous or maculopapular eruption with pruritus was documented on the admission exam, so the timestamp is 0 hours, since it happens right at admission. DRESS syndrome has no specific time, but it should happen soon after admission to the hospital, so we use our clinical judgment to give the diagnosis of DRESS syndrome the timestamp 0. then the output should look like

18 years old| 0
male | 0
admitted to the hospital | 0
fever | -72
rash | -72

```
    acne |  -672
    minocycline |  -672
    increased WBC count | 0
    eosinophilia| 0
    systemic involvement| 0
    diffuse erythematous or maculopapular eruption| 0
    pruritis | 0
    DRESS syndrome | 0
    fever persisted | 0
    rash persisted | 0
    discharged | 24
```
   Separate conjunctive phrases into its component events and assign them the same timestamp (for example, the separation of 'fever and rash' into 2 events: 'fever' and 'rash')  If the event has duration, assign the event time as the start of the time interval. Attempt to use the text span without modifications except 'history of' where applicable. Include all patient events, even if they appear in the discussion; do not omit any events; include termination/discontinuation events; include the pertinent negative findings, like 'no shortness of breath' and 'denies chest pain'.  Show the events and timestamps in rows, each row has two columns: one column for the event, the other column for the timestamp.  The time is a numeric value in hour unit. The two columns are separated by a pipe '|' as a bar-separated file. Skip the title of the table.
",
provided the body of the case report in {original_text}, and incorporated the initial output in {response} with {feedback_prompt} of "are you sure?".  The version GPT-4 w/feedback used this feedback loop twice, and we used the {response} as the final output.  We used a temperature of 0 for GPT-4 and GPT-4 w/feedback, and 1 for O1-preview per necessity.  Because GPT-4 and O1-preview have token limits of 8,192 and 32,768 respectively, we right-truncated the input of the prompt above to fit accordingly.

*Assessment*.
LLM annotations were assessed via comparison with manual annotation by a clinically-trained expert and for reliability via inter-LLM agreement.  The comparisons were conducted at the event level, where text spans were matched to one another per-report, and at the temporal level via ordering agreement (concordance) and relative time error (absolute time error, or absolute time discrepancy in the case of inter-LLM agreement).

To identify text matches a distance matrix was used between two annotated events lists of the same report.  Since the events were generated from the same report implying a 1-to-1 event correspondence, we adopted a recursive best match to reduce erroneous matches, *i.e.*, multiple events from the first list mapping to the same event in the second list.  Procedurally, if there were multiple events matching an event in the second list, the lowest distance match was selected, and those matched events were removed from the lists in the recursive call. To compute distance between a pair of text spans, we tested Levenshtein distance (the minimum number of insertion/deletion/substitutions) and cosine distance of sentence transformers mean embeddings from models "all-MiniLM-L6-v2" and "S-PubMedBert-MS-MARCO"[17,18].  Upon visual inspection of the matches on the development set, we determined that "S-PubMedBert-MS-MARCO" produced the best text span matches.  A cosine distance of 0.1 was chosen as the threshold for event match by manual review.

Among matched events, the relative times were compared for temporal ordering using concordance, which gives the empirical probability of correct ordering among all pairs of events (specifically, the probability of "not incorrect" ordering), where correct ordering is assumed to be given by the manual annotations as ground truth.  Concordance was calculated using the R glmnet Cindex function. The absolute time error or discrepancy was calculated between all matched events to quantitatively characterize the relative time annotations.  These temporal assessment measures were selected to highlight performance in terms of temporal ordering and time recovery, respectively.

For reference, the ten expert-annotated case files are: PMC4300884, PMC4478313, PMC4818304, PMC5667582, PMC6030904, PMC6034490, PMC7337692, PMC7747049, PMC8127753, and PMC9871993.  Code for annotation, assessment, and annotations is available at: https://github.com/jcweiss2/LLM-Timeline-PMOA/.

## Results

*Annotations*. Manual annotation results in 320 event-time pairs, with an average of 32 events annotated per report (Table 1). The average number of distinct relative times extracted is 6. The relative time distribution varies from time of presentation to multiple years from presentation, with 55% of events located at time of presentation. In the 10 case reports, none of the events were located at hourly resolution (Figure 2). In contrast to the manual annotations, LLM annotations have an average of 39–46 events per report (range: 122–144% of the manual annotations). Despite this difference, the automated annotation process identifies a similar number of distinct relative times: 6 to 8 (Table 1).

Table 1. Descriptive statistics of the manual and LLM annotations.

| Statistic | Manual[1] | GPT-4[1] | GPT-4 w/feedback[1] | O1-preview[1] |
|---|---|---|---|---|
| Events | 32 [14,70] | 46 [20,297] | 44 [16,93] | 39 [27,58] |
| Distinct Times | 6 [2,13] | 6 [1,16] | 6 [1,19] | 8 [3,13] |

[1] Mean [Minimum,Maximum]

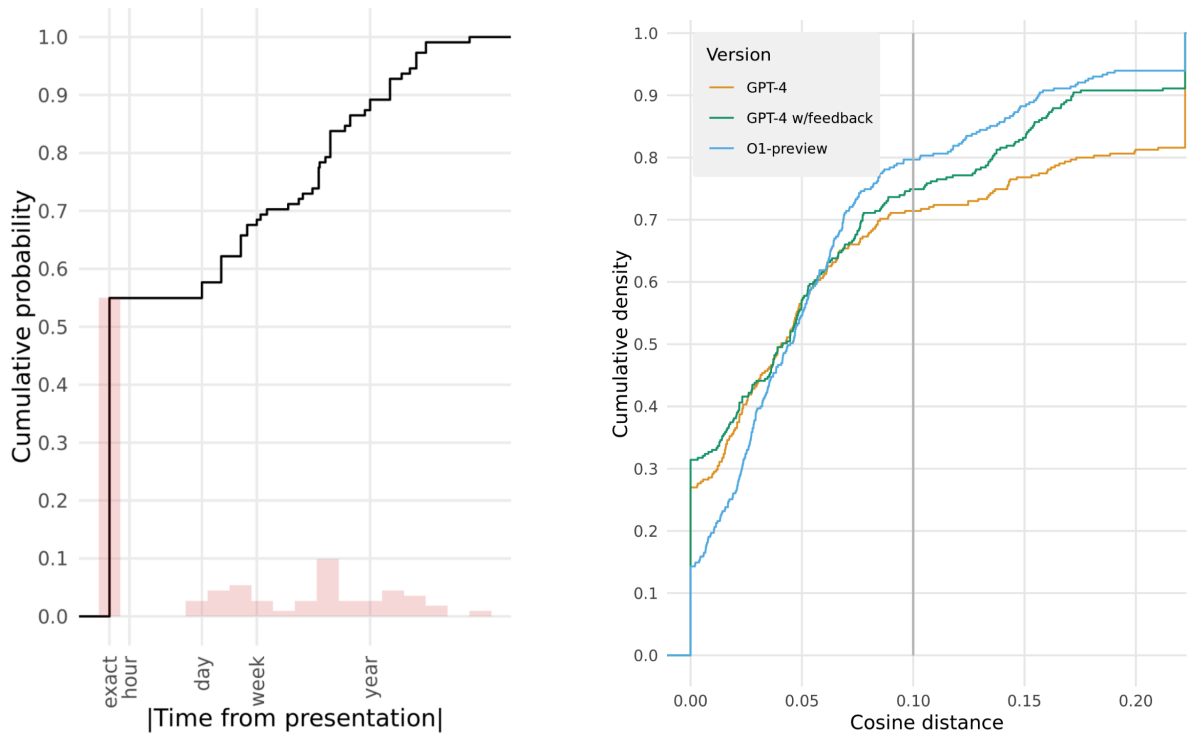

Figure 2. Distribution of absolute time relative to presentation of manual annotations (left), manual event match rate by LLM version (right).

*Event matching*.
While the manual annotation results in fewer events as the LLM annotations (Figure 2, left), the event match rate at a cosine distance threshold from 0.1 ranges between 70–80%. Modifying the threshold could lead to marginal increases in text matches, at the cost of increased false positives (Figure 2, right). Potential factors for the event recall could be the truncation from the token limit of GPT-4, and the incomplete splitting of conjunctive lists, resultings in multiple manually annotated events per event list. The O1-preview annotations have notably few exact textual matches, suggesting frequent rewording or rephrasing of the text span.

| file.name | Manual annotation | GPT-4 w/feedback | Cosine distance |
|---|---|---|---|
| PMC7337692.csv | Oxygenation improved | Oxygenation improved to requiring only a 2 L nasal cannula | 0.0757 |
| PMC6030904.csv | cholecystitis | Gallbladder duplication with cholecystitis suspected | 0.0777 |
| PMC6030904.csv | right gallbladder had purulent bile with thickening | Pathology reported two permeable cystic ducts, and two separ… | 0.0842 |
| PMC7747049.csv | psoriasis | diagnoses: hypothyroidism, insulin resistance, psoriasis, and hir… | 0.0888 |
| PMC6030904.csv | postoperative course of the patient was uneventful | Follow-up controls showing patient doing well | 0.0989 |
| PMC5667582.csv | skin ulcer with a cover of white pus moss | self-rupture of blister and formation of skin ulcer with pus | 0.1004 |
| PMC6030904.csv | second gallbladder appeared normal | Final diagnosis of Y-shaped gallbladder duplication with cholec… | 0.1047 |
| PMC6030904.csv | laparoscopy | Laparoscopy identifying gallbladder duplication | 0.1052 |
| PMC5667582.csv | amphotericin B | initiation of anti-Pneumocystis therapy | 0.1144 |
| PMC4818304.csv | difficulty was encountered due to the large cystic div… | ciliated foregut cyst of the gallbladder diagnosis | 0.1207 |
| PMC8127753.csv | Otoscopic examination | sinus tract was traced upwards towards the opening at the exte… | 0.1275 |

Figure 3. Potential event matches around the threshold of cosine distance = 0.1, ordered by cosine distance of the mean tokenized embedding from GPT-4 w/feedback. True matches (blue) tend to occur at smaller cosine distances and mismatches (white/gray) at larger distances.

*Temporal matching.*
Comparing the relative times of matched events with that of the manual annotations, we find that the LLM annotations possessed high concordance (means—GPT-4: 0.912, GPT-4 w/feedback: 0.876, and O1-preview: 0.951; boxplots—see Figure 4, left). Fifty to seventy-five percent of matched events have the same relative times as the manual annotation time, and approximately 70–85 percent of events have time errors within 24 hours (Figure 4, center). The distribution of time errors do not vary substantially across LLM versions. In subgroups of the manually-annotated relative times (relative to time of case presentation), we observe that the large time errors (greater than 1 week) tend to occur when the manual annotation time is away from the time of case presentation (Figure 4, right).

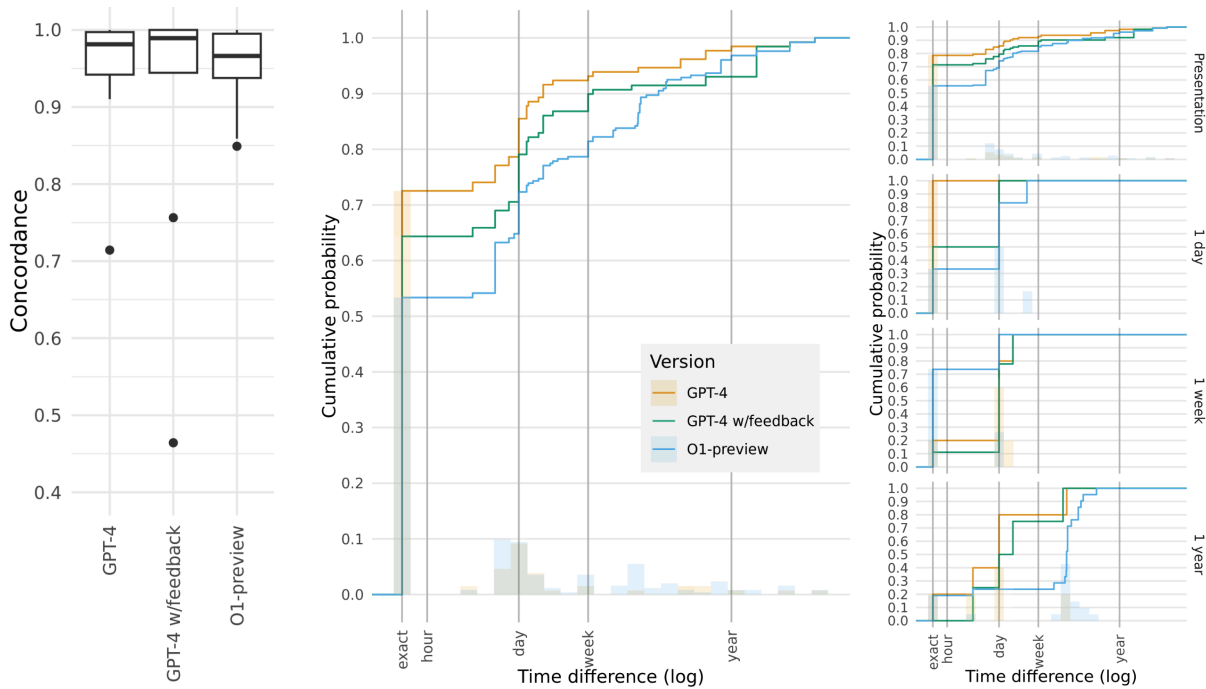

Figure 4. Temporal matching performance measures. Shown are concordance (left), absolute time error (center), and absolute time error by groups of time-from-presentation: at presentation, ≤1 day from presentation, ≤1 week from presentation, ≤1 year from presentation (right).

*Inter-LLM agreement.*
Annotation of the 93 case reports using the GPT-4 and GPT-4 w/feedback approaches result in 3,998 matched events. Of those, 3,103 (78%) find matching events with cosine distance less than 0.1. Among matched events, temporal ordering is high (mean: 0.97, boxplot—see Figure 5 left). Compared to manual annotations, the LLM annotations have similar levels of concordance, higher probability of identical time recoveries, and similar increased discrepancies as the time from case presentation increases.

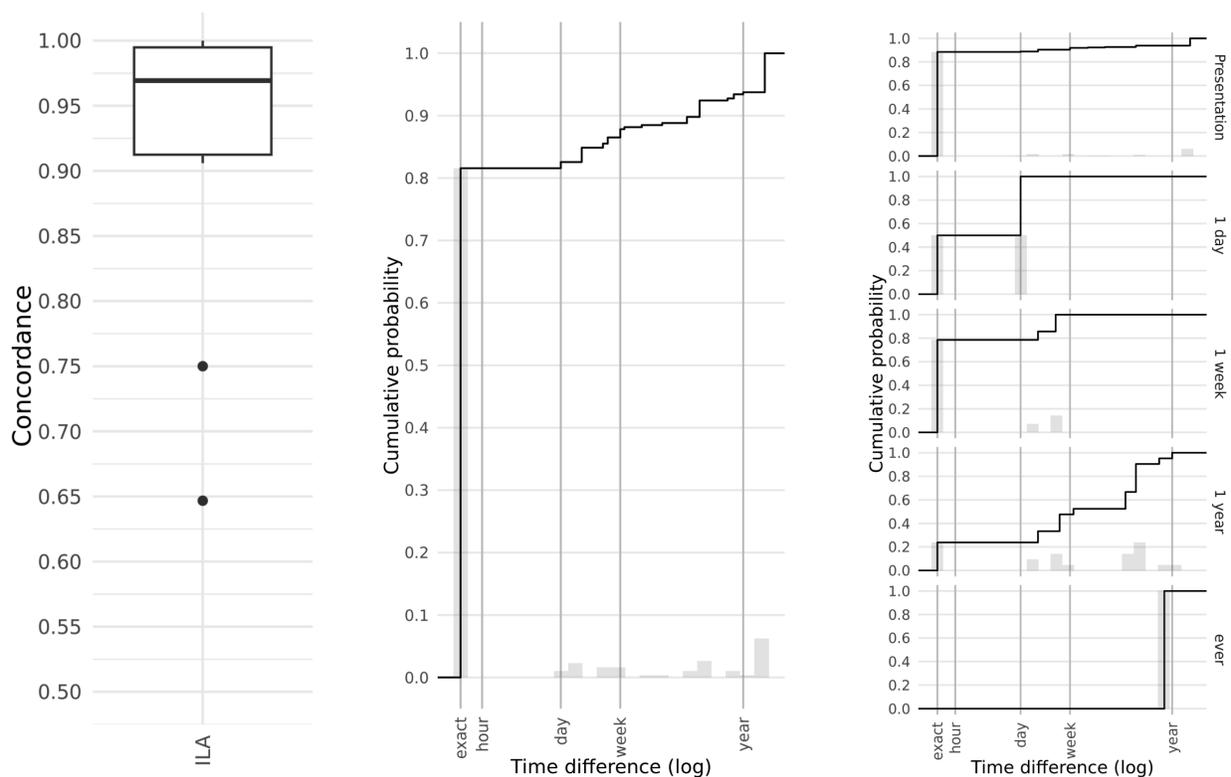

Figure 5. Inter-LLM agreement (ILA) between GPT-4 and GPT-4 w/feedback, measured by concordance (left), absolute time discrepancy (center) and absolute time discrepancy by subgroups of time to case presentation (right).

**Discussion**
Our results suggest that LLM annotation queries are able to effectively extract event-time tuples from free-text case reports. While event recall is modest (70–80%), the similar number of distinct times identified via LLM approaches suggests that the overall event timing structure is captured and that the determination of constituent events may be improved further (Table 1, Figure 4 left). While temporal ordering appears to be largely correct as seen in the high concordances, there remain discrepancies in relative time localization, with 10–20% of events being mislocated by at least 1 week and 2–7% of events mislocated by at least 1 year (Figure 4 center). This could have implications during alignment with complementary temporal data streams, where the timestamp errors can affect anchoring, feature availability for forecasting, right censorship, *etc*.

The inter-LLM agreement plots largely resemble the manual-LLM agreement plots (Figures 4 and 5), which assuages concerns about chance unrepresentativeness due to the small size of the manual annotation corpus. Additionally, the same trends are observed in the ILA plots—high concordance, similar discrepancy distributions, both in the full analysis and in subgroups of relative time—suggesting that ILA assessment could serve as a reasonable proxy for iterative improvement to reducing the dependence on high-cost expert annotation.

This work has several limitations; we describe them with respect to task specification, assessment design choices, and statistical analysis. First, we chose as part of our task to model events as minimally-modified text spans, which makes inverse mapping to referent passage in the text straightforward (an informal attempt to the LLM to maintain a character position backreference was hampered by confabulation). This contrasts with the approach of having events be named entities that could be mapped to existing clinical ontologies[4,19]. Our choice increases our reliance on textual comparison via tokenized embeddings, which required repeated review of event matches ordered by cosine distance. Our intention was that, by preserving access to the backreference, downstream applications such as forecasting and process discovery will be able to use the context during modeling.

Second, previous works have modeled a larger set of time expressions, the closest being timestamped intervals[7,8], and others expanding to time expressions of date, datetime, duration, frequency, and temporal relations. While in our initial exploration we considered using relative time intervals, we observed that the majority of events in our sample reports could be referenced with a single relative time, and that requesting additional temporal information often degraded query responses substantially. Future work could consider expansion of time points into intervals based on the relative times of our project, plus text span and context from the case report. Beyond time point to time interval concerns, there were also a number of instances where temporal ordering was specified without relative time specification that the expert annotator had difficulty annotating. For example, in "anticoagulation therapy [after surgery] was resumed once the patient had stabilized," the temporal relation is clear, however the expert would choose to select anticoagulation therapy as being initiated at the same time as stabilization and surgery, since surgical and stabilization durations were never specified. This example illustrates the limitation of single relative time specification, and could be expanded upon with temporal syntactic and probabilistic approaches[7,8].

Third, in our assessments we used recursive event matching to avoid many-to-1 mappings of the textual embeddings. This has the advantage of preventing large temporal discrepancies due to incorrect, low-distance event matches at the cost of lower event recall. When one event list contains conjunctive lists while the other splits them, the requirement of 1-to-1 matching prevents full matching. Since our emphasis is on getting high quality (if partial) time series, we argue that our emphasis on temporal consistency outweighs the lower recall, and that future work could focus on improving the regularity of list expansion beyond prompt engineering.

Fourth, we used only a sample of the PubMed Open Access case reports, which means that many case report subgroups remain un- or under-characterized. For example, none of the 10 reports that underwent manual review had events at the minute or hour resolution. These temporal resolutions may be important in fields such as critical care, anesthesiology, and procedural disciplines. Future work will involve expanding our annotation framework to the full PMOA case report corpus, and investigations across the diversity of case reports will be an important validation step and will help refine our annotation process. Additionally, biases that may be present in the LLMs may manifest in the query responses that could affect subgroup characterization and representativeness.

Additional approaches could expand on this work as future directions—including LLM annotation ensembling, prompt engineering, and active learning-based fine tuning—and could further improve the performance of the event-time extraction system. Downstream of our work, we anticipate that an event-time PMOA corpus could help improve forecasting models, process discovery, and AI-augmented clinical reasoning.

**Conclusion**
We find that LLM annotation processes can effectively recover ordering of clinical events from free text case reports as demonstrated by high concordance, while larger room for improvement remains in extracting complete event lists and recovering exact timing. Future work can expand the focus of LLM annotations to additional time expressions, develop enhanced processes for annotations using human-in-the-loop feedback, and conduct detailed assessments on focused clinical cohorts.

**Acknowledgments**
This research was supported by the Division of Intramural Research of the National Library of Medicine at the NIH.

**References**
1. Kohane IS. Temporal reasoning in medical expert systems. PhD Thesis, Boston University, 1987.
2. Zhou L, Hripcsak G. Temporal reasoning with medical data—a review with emphasis on medical natural language processing. *Journal of Biomedical Informatics*. 2007 Apr 1;40(2):183-202.
3. Pustejovsky J, Lee K, Bunt H, Romary L. ISO-TimeML: An international standard for semantic annotation. In *LREC,* 2010 May 18 (Vol. 10, pp. 394-397).

4. Sun W, Rumshisky A, Uzuner O. Evaluating temporal relations in clinical text: 2012 i2b2 challenge. *Journal of the American Medical Informatics Association*. 2013 Sep 1;20(5):806-13.
5. Cheng C, Weiss JC. Typed markers and context for clinical temporal relation extraction. In *Machine Learning for Healthcare Conference*, 2023 Dec 22 (pp. 94-109). PMLR.
6. Kougia V, Sedova A, Stephan A, Zaporojets K, Roth B. Analysing zero-shot temporal relation extraction on clinical notes using temporal consistency. *arXiv preprint* arXiv:2406.11486. 2024 Jun 17.
7. Leeuwenberg A, Moens MF. Towards extracting absolute event timelines from English clinical reports. *IEEE/ACM Transactions on Audio, Speech, and Language Processing*. 2020 Sep 28;28:2710-9.
8. Frattallone-Llado G, Kim J, Cheng C, Salazar D, Edakalavan S, Weiss JC. Using multimodal data to improve precision of inpatient event timelines. In *Pacific-Asia Conference on Knowledge Discovery and Data Mining* 2024 May 1 (pp. 322-334). Singapore: Springer Nature Singapore.
9. Singhal K, Tu T, Gottweis J, Sayres R, Wulczyn E, Hou L, Clark K, Pfohl S, Cole-Lewis H, Neal D, Schaekermann M. Towards expert-level medical question answering with large language models. *arXiv preprint* arXiv:2305.09617. 2023 May 16.
10. PMC open access subset [Internet]. Bethesda (MD): National Library of Medicine. 2003 - 2024 September 14. Available from https://pmc.ncbi.nlm.nih.gov/tools/openftlist/.
11. Xie Q, Chen Q, Chen A, Peng C, Hu Y, Lin F, Peng X, Huang J, Zhang J, Keloth V, He H. Me llama: foundation large language models for medical applications. *arXiv preprint* arXiv:2402.12749. 2024 Feb 20.
12. Li R, Wang X, Yu H. Two directions for clinical data generation with large language models: data-to-label and label-to-data. In *Proceedings of the Conference on Empirical Methods in Natural Language Processing*. Conference on Empirical Methods in Natural Language Processing 2023 Dec (Vol. 2023, p. 7129).
13. Knez T, Žitnik S. Multimodal learning for temporal relation extraction in clinical texts. *Journal of the American Medical Informatics Association*. 2024 Jun 1;31(6):1380-7.
14. Yuan C, Xie Q, Ananiadou S. Zero-shot temporal relation extraction with ChatGPT. In the *61st Annual Meeting of the Association for Computational Linguistics*. 2023 Jul.
15. Wei CH, Allot A, Lai PT, Leaman R, Tian S, Luo L, Jin Q, Wang Z, Chen Q, Lu Z. PubTator 3.0: an AI-powered literature resource for unlocking biomedical knowledge. *Nucleic Acids Research*. 2024 Apr 4.
16. Peng Y, Chen Q, Lu Z. An empirical study of multi-task learning on BERT for biomedical text mining. *BioNLP 2020*. 2020 Jul 9:205
17. Wang W, Wei F, Dong L, Bao H, Yang N, Zhou M. MiniLM: deep self-attention distillation for task-agnostic compression of pre-trained transformers. *Advances in Neural Information Processing Systems*. 2020;33:5776-88.
18. Deka PR, Jurek-Loughrey AN, Padmanabhan D. Improved methods to aid unsupervised evidence-based fact checking for online health news. *Journal of Data Intelligence*. 2022 Nov;3(4):474-505.
19. Hu Y, Chen Q, Du J, Peng X, Keloth VK, Zuo X, Zhou Y, Li Z, Jiang X, Lu Z, Roberts K. Improving large language models for clinical named entity recognition via prompt engineering. *Journal of the American Medical Informatics Association*. 2024 Jan 27.